\documentclass[conference]{IEEEtran}
\IEEEoverridecommandlockouts
\usepackage{cite}
\usepackage{amsmath,amssymb,amsfonts}
\usepackage{algorithmic}
\usepackage{graphicx}
\usepackage{textcomp}
\usepackage{xcolor}
\usepackage{hyperref}
\usepackage{subfigure}
\usepackage{booktabs} 
\usepackage{tabularx}
\usepackage{float}
\usepackage{microtype}
\usepackage{color, colortbl}
\usepackage[ruled,vlined]{algorithm2e}
\usepackage{url}
\usepackage{algorithmic}
\usepackage{mathtools}
\usepackage{dblfloatfix}
\usepackage{fourier} 
\usepackage{array}
\usepackage{makecell}
\usepackage{multirow}
\usepackage{color, colortbl}
\usepackage{orcidlink}

\usepackage{amssymb}
\usepackage{pifont}
\newcommand{\cmark}{\ding{51}}%
\newcommand{\xmark}{\ding{55}}%
\definecolor{Gray}{gray}{0.9}
\def\BibTeX{{\rm B\kern-.05em{\sc i\kern-.025em b}\kern-.08em
		T\kern-.1667em\lower.7ex\hbox{E}\kern-.125emX}}
\begin{document}
	
	\title{LEAPMood: Light and Efficient Architecture to Predict Mood with Genetic Algorithm driven Hyperparameter Tuning}
	\author{\IEEEauthorblockN{Harichandana B S S}
		\IEEEauthorblockA{\textit{Samsung R\&D Institute} \\
			Bangalore, India \\
			hari.ss@samsung.com \\
			\orcidlink{0000-0002-6123-2249} 0000-0002-6123-2249}
		\and
		\IEEEauthorblockN{Sumit Kumar}
		\IEEEauthorblockA{\textit{Samsung R\&D Institute} \\
			Bangalore, India \\
			sumit.kr@samsung.com \\
			\orcidlink{0000-0003-0373-1397} 0000-0003-0373-1397
		}
		
	}
	
	\maketitle
	
	\begin{abstract}
		Accurate and automatic detection of mood serves as a building block for use cases like user profiling which in turn power applications such as advertising, recommendation systems, and many more. One primary source indicative of an individual's mood is textual data. While there has been extensive research on emotion recognition, the field of mood prediction has been barely explored. In addition, very little work is done in the area of on-device inferencing, which is highly important from the user privacy point of view. In this paper, we propose for the first time, an on-device deep learning approach for mood prediction from textual data, LEAPMood. We use a novel on-device deployment-focused objective function for hyperparameter tuning based on the Genetic Algorithm (GA) and optimize the parameters concerning both performance and size. LEAPMood consists of Emotion Recognition in Conversion (ERC) as the first building block followed by mood prediction using K-means clustering. We show that using a combination of character embedding, phonetic hashing, and attention along with Conditional Random Fields (CRF), results in a performance closely comparable to that of the current State-Of-the-Art with a significant reduction in model size (> 90\%) for the task of ERC. We achieve a Micro F1 score of 62.05\% with a memory footprint of a mere 1.67MB on the DailyDialog dataset. Furthermore, we curate a dataset for the task of mood prediction achieving a Macro F1-score of 72.12\% with LEAPMood.
	\end{abstract}
	
	
	\section{Introduction}
	
	Mood changes are experienced by everybody. Recognizing and keeping track of these moods, can improve the overall experience for the user during any human-computer interactions and additionally help improve their mental wellbeing. Though the benefits of accurately predicting the user's mood state are numerous, doing so is a very complex task \cite{mikus2018predicting}. This is because moods can be affected by various factors being psychological state, environmental changes, etc.
	
	\begin{figure}[ht]
		\centering
		\includegraphics[width=0.5\linewidth]{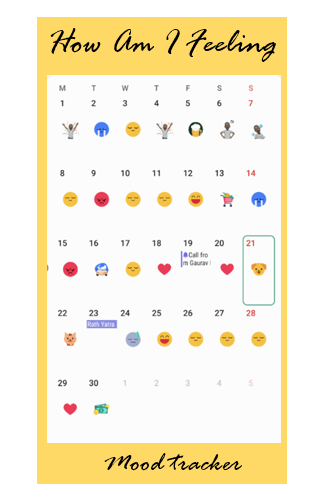}
		\caption{Mood Calendar: Automatically keep track of user's daily mood states}
		\label{fig:intro}
		\vskip -0.15in
	\end{figure} 
	
	Technology has a huge impact on the emotional state of a person nowadays, due to growing user engagement in social media. The latest data show that 57\% of the total global population use social media around the world in July 2021. These figures suggest that more than 9 in 10 internet users now use social media each month \cite{data}. As a direct outcome, one major indicator of a user's mood is the textual content accessed or created by the user. 
	
	Detection of mood becomes more imperative in post covid world. At large, all of the studies on the psychological disorders during the COVID-19 pandemic have reported that the affected individuals show several symptoms of mental trauma, such as emotional distress, depression, mood swings, insomnia, anger, etc. \cite{salari2020prevalence}. In this paper, we focus on automatically predicting the mood of an individual from textual data. This can serve many use cases like a mood calendar to help track one's state of mood as shown in Figure~\ref{fig:intro}.

	The first requirement for this problem statement is to ensure the privacy of user data. Thus, we aim to create an effective and lightweight solution that is deployed completely on-device to protect the user's privacy. Today there are over six billion smartphone subscribers worldwide and is forecasted to further grow by several hundred million in the next few years \cite{statistica}. A prediction by  Gartner shows that 80\% of smartphones will have on-device AI, and this is up from just 10\% in 2017 \cite{gartner}. This proves that on-device solutions for such problem statements are very crucial.
	
	While there has been extensive work done in the field of Emotion recognition, mood prediction is relatively untouched. The concepts of mood and emotion may seem similar but they consist of major differences in reality. Mood can last longer than emotions \cite{moodvsemotion} and are very subtle compared to emotions which can be intense like fear, joy, etc. and this may result in unawareness of one's mood i.e. good or bad until self-reflection. Mood tracking can be done at set time intervals, to identify patterns in how their mood varies. It is found that Emotions come first, then moods develop from a combination of feelings \cite{emotionasmood}. Hence, we can use emotions detected and predict mood for a specific time duration.
	
	To this end, we propose a novel Light and Efficient Architecture to Predict Mood (LEAPMood). To summarize our major contributions in this paper:
	\begin{itemize}
		\item We propose a novel pipeline LEAPMood, for efficient on-device mood prediction from textual data. 
		\item We propose a novel lightweight model for on-device ERC as a subcomponent in our pipeline and introduce phonetic hashing as a pre-processing step to capture Out of Vocabulary (OOV) and exaggerated emotions.
		\item We introduce a new Fitness function for Hyperparameter tuning using GA focused on on-device deployment.
		\item We benchmark our solution for on-device ERC on the DailyDialog dataset and analyze the overall performance of LEAPMood on a custom dataset for mood prediction.
	\end{itemize}

	\section{Related Works}
	
	Psychologists have attempted to understand and explain human emotion since the 19th-century \cite{thanapattheerakul2018emotion}. The concept of Affective computing was proposed and studied as early as the 1990s where R.W. Picard talks about this concept in his book \cite{picard2000affective}. Soon, there has been numerous and extensive work done in this field attracting many researchers to detect an individual’s emotional state from various modalities as sources such as eye gaze, audio, gestures, etc. \cite{10.1145/306668.306683}\cite{840655}\cite{ inproceedings }\cite{nguyen}\cite{499429}. One such popular modality to recognize human emotions is text. Emotion Recognition in conversation (ERC) is challenging in the sense that the same set of words or phrases can be used to depict different emotions.
	
	Early research in the field of ERC involved keyword-based techniques which are one of the most intuitive and naive approaches. \cite{wilson2004just} use syntactic features and identify opinions in deeply nested clauses to classify their strengths. Following this, some semi-automatic methods for creating emotion dictionaries utilizing methods like WordNet were explored \cite{bracewell2008semi}. In the following years, limitations of keyword-based techniques were uncovered by researchers like \cite{wu2006emotion}. 
	
	Learning-based approaches have been explored to overcome the limitations of keyword-based methods. \cite{4368008} propose a set theory and Support Vector Machine (SVM) based approach. This was followed by the use of Conditional Random Fields (CRF) which showed improved performance for emotion classification \cite{yang2007emotion}.
	
	In the coming years, RNN based architectures for multi-modal emotion architectures \cite{zadeh2017tensor}. DialogueRNN introduced in \cite{majumder2019dialoguernn} used GRUs to model global and speaker states. With the advance of Transformer based models, many works have been done using different variants of these \cite{shen2020dialogxl}\cite{lee2021graph}. The current SOA CESTa uses a transformer-based encoder along with CRF \cite{wang2020contextualized}. One major drawback is that these models are very huge in size making them not suitable for on-device deployment.
	
	In contrast to emotion recognition, mood prediction has been explored mainly from the biological sensor-based approach \cite{jaques2017multimodal}\cite{cho2019mood}. \cite{liu2020multimodal}\cite{roshanaei2015features} attempt to predict mood from textual and other mobile data but do not consider fine-grained information like emotional states experienced by user to do so. We attempt to do this for the first time.

	Since we require a solution completely on-device, it is crucial to get the most out of small model architecture. Hyperparameter tuning has proved to be effective to do so. Research in this field include \cite{falkner2018bohb}\cite{akiba2019optuna} and many more. We observe that there is no work done in this field specifically for on-device deployment. Thus, we explore this area. 
	
	\begin{figure}[b]
		\centering
		\includegraphics[width=\linewidth]{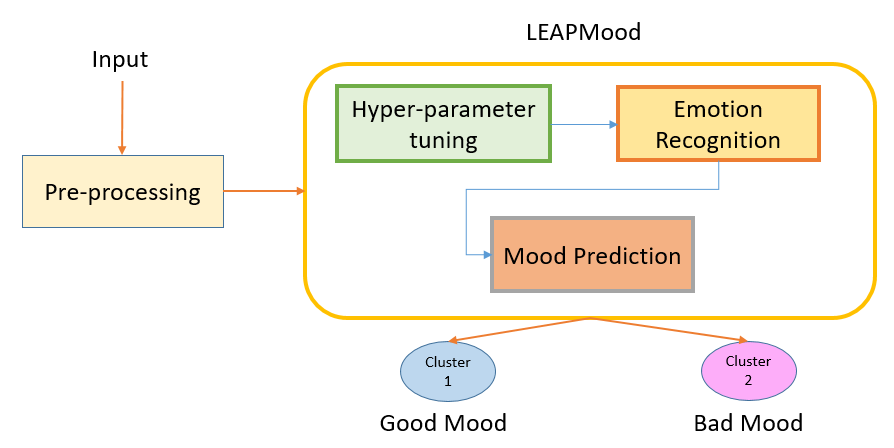}
		\caption{Overall Pipeline of LEAPMood}
		\label{fig:pipeline}
	\end{figure}

	\section{Pipeline} \label{s:pipeline}
	
	The proposed pipeline consists of first identifying the emotion triggered by a specific text and then, based on the duration on which mood is to be determined, processing the emotions recognized to predict the mood. The first challenge encountered for modeling this solution is the lack of a dataset that supports mood prediction. 
	
	On the other hand, there is extensive work done in the field of sentiment analysis and emotion recognition using textual input. We use DailyDialog dataset \cite{li2017dailydialog} to train and test our model on the emotion recognition task. We notice one disadvantage of such open datasets which is that they do not give complete insight on the real-world performance due to lack of coverage on spell mistakes, short forms, exaggerated words, etc. which may lead to many unknown tokens after pre-processing which in turn results in loss of important information to give correct results. 
	
	To tackle such scenarios, we use our custom data as well which consists of group chat data extracted and annotated on both emotions and mood categories. This helps further to train and test our complete pipeline for this proposed solution which is shown in Fig.~\ref{fig:pipeline}. It is worth noting that we use both character and word embedding along with phonetic hashing to reduce OOVs. The detailed explanations for each module are covered in the following subsections.

	\begin{table}[t]
		\caption{DailyDialog dataset statistics}
		\label{t4}
		\vskip -0.2in
		\begin{center}
			\resizebox{\columnwidth}{!}{
				\begin{tabular}{c c c c c c c c}
					\toprule
					\textbf{Emotion} & \textbf{Anger} & \textbf{Disgust} & \textbf{Fear} & \textbf{Happy} & \textbf{Sad} & \textbf{Surprise} & \textbf{Other}  \\
					\midrule
					\textbf{\#} &  1022& 353&74& 12885&1150& 1823&85572 \\

					\bottomrule
				\end{tabular}
			}
		\end{center}
		\vskip -0.15in
	\end{table}
	
	\begin{figure}[t]
		\centering
		\includegraphics[width=0.7\linewidth]{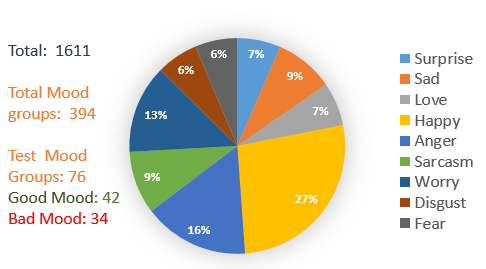}
		\caption{Custom Dataset Distribution}
		\label{fig:dataset_custom}
		\vskip -0.1in
	\end{figure}
	
	\begin{table}[t]
		\caption{Custom Dataset Samples sorted according to Timestamps.}
		\label{t:customdata}
		\begin{center}
			\begin{tabular}{m{0.8\columnwidth} m{0.1\columnwidth}}
				\toprule
				\textbf{Dialogue} & \textbf{Emotion}  \\
				\midrule
				yay! tmrws a holiday  & Happy  \\
				ill see you at the movie in an hr &  Happy\\
				brooo..i crashed my fathers car. i don't know what to do & Fear \\
				he hit my car and ran away & Anger \\
				\midrule
				Overall Mood & Bad \\
				\bottomrule
			\end{tabular}
		\end{center}
		\vskip -0.1in
	\end{table}
	\begin{figure*}[t]
		\centering
		\includegraphics[width=0.9\linewidth]{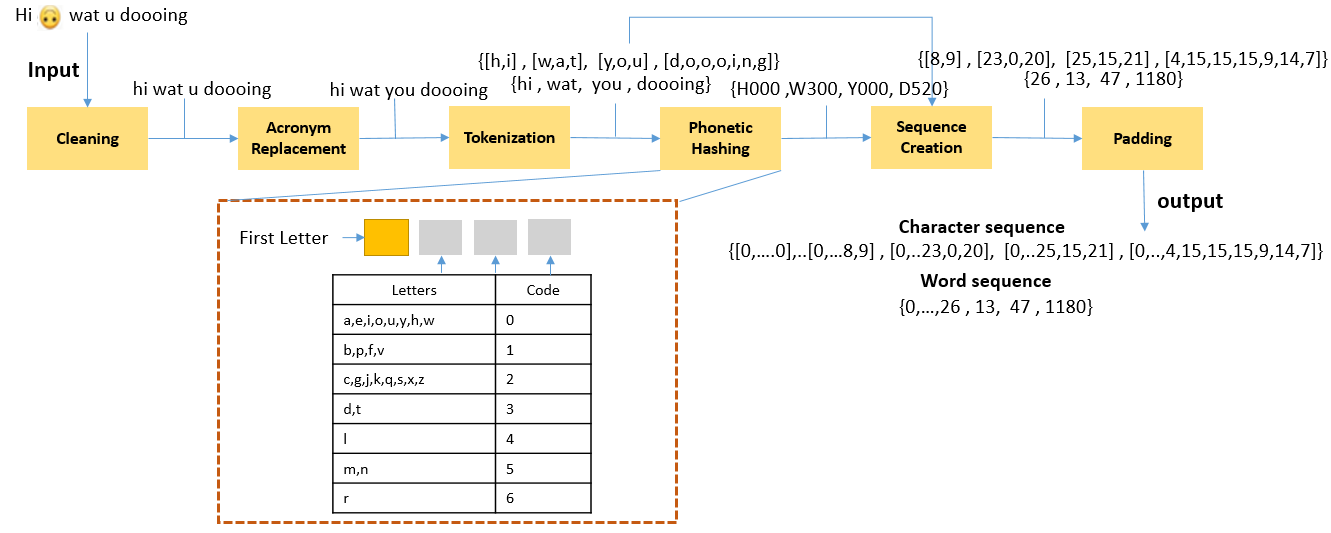}
		\caption{Preprocessing Pipeline.}
		\label{fig:preprocess}
		\vskip -0.1in
	\end{figure*}
	\subsection{Dataset} \label{s:dataset}
	We use the DailyDialog dataset \cite{li2017dailydialog} to train and benchmark our model for the task of emotion recognition. The dataset consists of 7 emotion categories, details of which can be seen in Table \ref{t4}. It can be observed that the data is highly imbalanced. To solve this issue, we use class weights while training the model which is explained in section \ref{s:model}.

	We create our custom dataset to test the real-time performance of the model. We curated the data by extracting multiple group chats from WhatsApp and removing the private content of the involved individuals. We then annotate the data according to the emotion from the textual content collecting over 1600 samples. This is done by 10 annotators from diverse age and gender groups.  The details of our dataset are shown in Fig.~\ref{fig:dataset_custom}. The data is further pre-processed before model training as explained in Section~\ref{s:preprocessing}.
	
	To further train our solution for mood prediction, we utilize the output logits from our ERC module. Since mood is an aggregation of various emotions over time, we group our samples according to timestamps and annotate these groups using binary labels indicating the mood i.e Good Mood or Bad Mood.
	We get a total train set of 394 and a test set of 76 mood groups as shown in Fig.~\ref{fig:dataset_custom}. One sample of such grouping is shown in Table~\ref{t:customdata}. To ensure the correctness of these labels, we cross-verify the labels with help of the above-mentioned annotators. The details of mood prediction are explained in the following Section \ref{s:model}.

	\subsection{Pre-processing} \label{s:preprocessing}
	The pre-processing of the data involves the following steps: Cleaning, Replace acronyms, Tokenization, Phonetic hashing, Create Sequences, Padding.

	We begin by cleaning the input text sample which includes converting to lowercase, removing non-alphanumeric symbols using regex, removing emoji(s) if present (in the case of the custom dataset) and finally, replacing numbers with a number tag. After this cleaning process, we replace the acronyms. For this, we create a map of the 100 most frequently used acronyms curated from our custom dataset and replace them with their respective full forms.
	
	The next step in the pre-processing pipeline is tokenization. During this process, we also create a word vocabulary of size 30,000 tokens. This is done by first keeping track of the frequencies of occurrence of each token and choosing the top 30,000 most frequently occurring tokens. Along with this, we create a character vocabulary as well to support character embeddings. We create a character vocabulary of size 30 characters which includes all english alphabets and a handful of other frequently used characters/symbols. 
	
	\begin{table}[t]
		\caption{ Soundex: codes for phonetically similar words.}
		\label{t:soundex}
		\begin{center}
			\begin{tabular}{  c c | c c }
				\toprule
				\textbf{Word} & \textbf{Soundex code} & \textbf{Word} & \textbf{Soundex code} \\
				\hline
				Happyyyyyyy & H100 & Happy & H100 \\
				\hline
				Elephant  & E415 & Elefant & E415 \\
				\hline
				Awesome & A250 & Awesoooomeeee & A250 \\
				\hline
			\end{tabular}
		\end{center}
		\vskip -0.2in
	\end{table}
	Following this step, we use phonetic hashing of tokens. This is done using the Soundex algorithm \cite{beider2010phonetic}. This results in a four-character code for any token with the characteristic of having the same codes for phonetically similar words as shown in Fig.~\ref{fig:preprocess}. Due to this property, this is a possible solution for spell errors and exaggerated words, as shown in Table~\ref{t:soundex}. For this, we map the phonetic codes for tokens in vocabulary, and during tokenization, in the case of an OOV, we match its phonetic code with that of the ones in the vocabulary. If a match is found,  we replace the token with the one in the vocabulary. Thus, we avoid tagging these as unknown and risk loss of important information.
	
	Next, we create input sequences by mapping the tokens and characters with their respective IDs in the word and character vocabularies created in the previous steps. This is then followed by padding the sequences to ensure uniform input size across all samples for model training.
	
	The complete pipeline is shown in Fig.~\ref{fig:preprocess}. 
	The final pre-processed samples are then fed to our model for training.

	\subsection{LEAPMood} \label{s:model}
	We propose a novel architecture LEAPMood, for efficient on-device mood prediction from textual data. As mood can be determined by the set of emotions experienced during the required period, rather than training directly for mood prediction, we model the solution as two components:   
	This consists of two main components: 
	
	\begin{figure*}[t]
		\centering
		\includegraphics[width=0.9\linewidth]{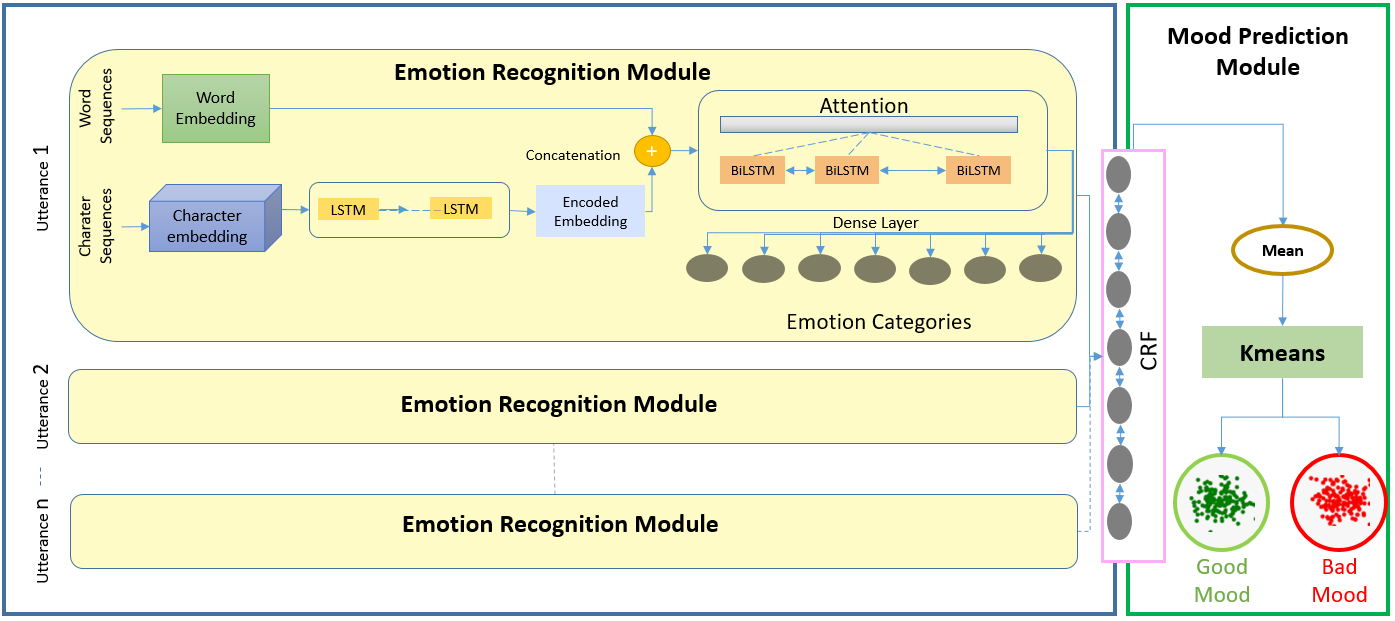}
		\caption{LEAPMood Architecture.}
		\label{fig:model}
		\vskip -0.1in
	\end{figure*}

	\begin{itemize}
		\item Emotion Recognition 
		\item Mood Prediction
	\end{itemize} 
	This also enables the use of ERC as an individual module for other use cases. The complete architecture of LEAPMood is shown in Fig.~\ref{fig:model}. 
	\subsubsection{Emotion Recognition in Conversation}\label{s:ERC}
	The first component of ERC mainly consists of five prominent layers: Word Embedding, Char Embedding, Long Short-Term Memory (LSTM) encoder for char embeddings, Bidirectional LSTM (BILSTM), Attention, and CRF as shown in Fig.~\ref{fig:model}.
	
	The pre-processed dataset having word sequences after phonetic hashing (Input-1) and character sequences (Input-2) along with one-hot encoded ground-truth labels are fed to the model. The input shapes are set to (,100) and (,100,10) respectively where the maximum sequence length is set to 100, and the maximum character sequence length to 10. The first layer is the word embedding layer which uses Input-1 and the embedding size is set to 56. Following this, we define a TimeDistributed Layer with Embedding layer which is applied on Input-2, resulting in char embeddings as output. For this, we set the character embedding dimension to be 16. We then use an LSTM layer with 20 units to encode these character embeddings and make them compatible to concatenate with the word embeddings. After concatenating both word and character embeddings, we use a Bidirectional LSTM with 57 units followed by a Dense Layer with softmax activation to get the logits for the input sample. This is followed by a CRF Layer to model the contextual information of previous and future labels. While compiling the model, we use Adam optimizer and cross-categorical entropy loss with class weights calculated to balance the loss on different imbalanced class labels. We train for 25 epochs with EarlyStopping, batch size of 90, and a learning rate of 0.0001. 
	
	These parameters along with other model hyperparameters are determined using a unique and novel Genetic Algorithm based hyperparameter optimization which will be discussed in detail in Section \ref{s:HP}.

	\subsubsection{Mood Prediction}\label{s:mood prediction}
	Following the emotion recognition module, we use the K-means clustering algorithm to predict the mood. The mood of a group of textual data can be predicted from the individual emotions recognized from the previous component as explained in Section~\ref{s:model}. We use clustering because we observe that for any positive emotion, the logits for all the positive emotion categories are high and this applies similarly for negative emotions. Thus, clustering will be able to group these and also does not require much memory space.
	
	We use the prepared dataset where we group batches of textual data of similar timestamps. We create such batches of text grouping data within the window of a 60-minute threshold. This configuration enables to predict mood for an interval of 1 hour for the user. This is configurable according to the user's needs. Following the creation of these batches of text, we take the aggregate of the logits corresponding to each class label according to \eqref{eq:Kmeans}.
	\begin{center}
		\vskip -0.2in
		\begin{equation}
		\label{eq:Kmeans}
		Aggregate\_logits = \big\{\frac{\sum_{i=1}^{N} x_{i1}}{N}, \frac{\sum_{i=1}^{N} x_{i2}}{N}, ..., \frac{\sum_{i=1}^{N} x_{iM}}{N}\big\}
		\end{equation}
	\end{center}
	Where N is the number of textual data samples in the group and M is the number of labels. $x_{iM}$ is the value of the $i^{th}$ sample for the $M^{th} label$. Finally, we use these logits to perform K-means clustering. We choose the value of K to be 2, giving two clusters one representing a good mood and the other as a bad mood. We choose to first experiment with a small number of clusters to avoid wrong predictions. Going forward, we will explore using higher values of K and thus predict fine-level mood other than just good or bad.

	\subsection{Hyperparameter tuning} \label{s:HP}
	\begin{figure}[ht]
		\centering
		\includegraphics[width=0.8\linewidth]{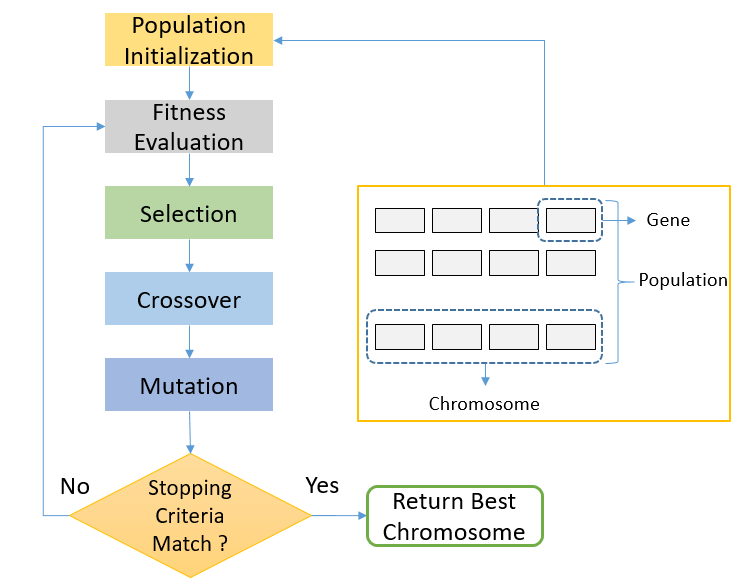}
		\caption{Genetic Algorithm}
		\label{fig:ga}
		\vskip -0.1in
	\end{figure}
	We use the Genetic Algorithm for Hyperparameter tuning. Since we focus on the on-device deployment of our model, we need to optimize both model performance and size. To do this, we use a novel fitness function that accounts for these two factors. The steps for on-device focused Hyperparameter tuning (as shown in Fig.~\ref{fig:ga}) are:
	\begin{itemize}
		\item Fitness Function and Stopping criteria definition 
		\item Population Initialization
		\item Fitness Evaluation
		\item Selection
		\item Cross-over
		\item Mutation
		\item Repeat till stopping criteria 
	\end{itemize}
	
	\subsubsection{Fitness Function and Stopping Criteria defination}
	Fitness function is defined to judge how good the individual solution is to solve the problem. In our case, it shows how good the choice of parameters is for the model performance. Additionally, we use the model size information corresponding to the chosen parameters to make sure we get the most optimal set of parameters that give the best performance with the least model size possible. We calculate the model size using the number of model weights and parameters required for the individual solution using \eqref{eq:ModelParams}.

	\begin{center}
		
		\begin{equation}
		\label{eq:ModelParams}
		\begin{multlined}[\columnwidth]
		HP_{i} = \left\{p_{i11}, p_{i21}, ...p_{ijk} \right\} where~p_{ijk} \in C_{k} \\
		Total\_Model\_Params_{i} = \sum \#C_{k}\big(p_ijk\big)~~ \forall p_{ijk} \in HP_{i}
		\end{multlined}
		\end{equation}
	\end{center}
	
	Here, $HP_{i}$ is the $i^{th}$ solution for Hyper parameters required to be tuned.  Each solution $HP_{i}$ contains set of hyper parameters $p_{ijk}$ representing $j^{th}$ parameter of component $k$, $C_{k}$. The total model parameters are calculated by summing up the number of weights required for each Component using parameter $p_{ijk}$ represented by $\#C_{k}\big(p_{ijk}\big)$.
	The final Fitness function used for our Model tuning is as represented by \eqref{eq:Fitness}.
	
	\begin{center}
		\vskip -0.2in
		\begin{equation}
		\label{eq:Fitness}
		Fitness_{i} = \frac{Model\_Accuracy_{i}\big(HP_{i}\big)}{Total\_Model\_Params_{i}}
		\end{equation}
	\end{center}
	
	Where $Model\_Accuracy_{i}\big(HP_{i}\big)$ is the Model accuracy after substituting parameters with those in $HP_{i}$. 
	
	\subsubsection{Population Initialization}
	
	Population in genetic algorithm consists of chromosomes which in turn consist of genes. In this case, each chromosome represents one possible solution ($HP_{i}$ in \eqref{eq:ModelParams}) and each gene represents a hyperparameter to be optimized i.e $p_{ijk}$. Each gene belongs to one of the components of the model architecture defined in Section~\ref{s:ERC} Ex: BiLSTM (represented by $C_{k}$). We optimize 10 hyperparameters namely: batch size, epochs, word embedding dimension, char embedding dimension, char LSTM hidden size, spacial dropout, LSTM dropout, LSTM recurrent dropout, BiLSTM hidden size, and BiLSTM recurrent dropout. These constitute the genes in each chromosome. We keep track of the type of gene i.e. continuous or discrete and set a range of [minimum, maximum] for each gene. We initialize these values randomly depending on the type and range. The population size is set to 7. This means we explore 7 different optimizations at a time.   
	
	The Stopping criteria determine when to stop executing the steps and return the most optimal solution at the time of termination. We set this to be the count of process loops reaching 250. That is, as soon as the process repeats 250 times, it terminates.

	\subsubsection{Fitness Evaluation}
	Following population initialization, we evaluate the fitness of each chromosome using our proposed fitness function. We train the model with the parameter values in the corresponding chromosome to get its fitness score.
	
	\subsubsection{Selection}
	We use Roulette wheel method for selection. In this, we divide the wheel into $N$ divisions where $N$ is equal to the population size which here is 7. Each division occupies an area proportionate to the fitness score of the corresponding chromosome according to \eqref{eq:roulettewheel}. 
	\begin{center}
		\vskip -0.2in
		\begin{equation}
		\label{eq:roulettewheel}
		Area_{i} = \frac{Fitness_{i}}{\sum_{i=1}^{N} Fitness_{i}}
		\end{equation}
	\end{center}
	
	This ensures that the best fit chromosomes have a high probability to get selected as parent chromosomes.

	\subsubsection{Cross-over}
	Cross over is used to recombine genes from two different parents to get a new offspring with its genes a mixture of the two selected parents. We use one point crossover technique where a point in the range of (0, Chromosome Size) is randomly picked called `crossover point'. After this, the genes to the right of the cross-over point are swapped resulting in two new offspring. This operation is repeated till a particular limit which is determined by `Crossover\_Rate'. We set the crossover rate as 0.5. This indicates that this process is repeated for $\big(Population\_Size * Crossover\_Rate\big)$ times.  Here chromosome size is 10. We randomly select two parents from the selected population and perform this operation.  
	
	\subsubsection{Mutation}
	Mutation is used to ensure genetic diversity in the offspring. While crossover interchanges the genes in parents, the individual values of the genes themselves are not changed. This is taken care of by using mutation.
	To do this we randomly choose a gene from a chromosome in the population after crossover, choose a value within its range defined earlier, and replace it with the new value. This is repeated for each chromosome and the number of genes chosen for mutation is equal to the ($Chromosome\_Size * Mutation\_Rate$). We set the mutation rate to be 0.25.

	\subsubsection{Repeat till Stopping Criteria}
	On execution of the above steps, we re-evaluate the fitness scores for all the new offspring in the population. We then check if the stopping criteria defined earlier is met. In our case, it is a count of process loops that is set to 250. If this is not yet met, we repeat the process.

	\subsection{On-device deployment}
	For deploying on-device, We also performed quantization using TensorFlow tflite as well on this model to further reduce the size which results in a model size of $\sim$1.9MB. We implement pre-processing techniques explained in section \ref{s:preprocessing} on Android. We then create a test application to run the model for on-device inferencing. This application takes input a set of textual input and outputs their respective emotion categories and also the overall mood.
	
	\begin{table}[ht]
		\caption{Model Performance on DailyDialog dataset}
		\label{tab:dailydialog}
		\centering
		\begin{tabular}{ c    c c  c }
			\toprule
			\textbf{Emotion} & \textbf{Precision} & \textbf{Recall}  & \textbf{F1 Score}  \\
			\midrule
			Anger     & 51.04    & 41.52  & 45.79         \\
			Disgust   & 16.21  & 38.29  &  22.78          \\
			Fear  & 19.51 & 47.05 & 27.58 \\ 
			Happiness  &  61.05 & 56.86 & 58.88  \\
			Sadness &  61.05 & 56.86 & 58.88 \\
			Surprise   & 42.77 & 63.79 & 51.21 \\
			\midrule
			Averaging (Macro) & 43.38 &  52.84 & 45.98\\
			\textbf{Averaging (Micro)} & \textbf{59.75} & \textbf{64.55}& \textbf{62.05} \\
			
			\bottomrule
		\end{tabular}
		\vskip -0.1in
	\end{table}
	
	\begin{table}[ht]
		\caption{Model Performance comparision with the state of the art and other baseline models for DailyDialog dataset}
		\label{tab:soa}
		\centering
		\begin{tabular}{ c    c c   }
			\toprule
			\textbf{Model} & \textbf{Model Size} & \textbf{Micro F1-Score}    \\
			\midrule
			DialogXL \cite{shen2020dialogxl}   & > 400MB  & 54.93    \\
			TUCORE-GCN \cite{lee2021graph}&    > 400MB & 61.91          \\
			\rowcolor{Gray}
			LEAPMood  &  \textbf{1.678MB} & \textbf{62.05}   \\      
			CESTa \cite{wang2020contextualized}  & > 200MB & \textbf{63.12}  \\

			\bottomrule
		\end{tabular}
		\vskip -0.1in
	\end{table}

	\begin{table}[t]
		\caption{Model Performance on Custom Dataset}
		\label{tab:custom data}
		\centering
		\begin{tabular}{   c  c c  c }
			\toprule
			Emotion          & Precision       & Recall        & F1 Score       \\ \midrule
			Surprise   &    60.86       &   43.75      &      50.90     \\
			Sad  &         47.72    &      60.00    &       53.16    \\
			Love    & 37.03 & 45.45  & 40.81 \\ 
			Happiness    & 48.68  & 72.54  & 58.26 \\
			Anger  &  62.96 & 36.69  & 46.36 \\
			Sarcasm    & 55.00 & 42.30 & 47.82 \\
			Disgust    & 54.16 & 35.13 & 42.617 \\
			Fear &  43.58 & 56.67 & 49.27 \\
			Worry  &  55.26 & 53.84 &  54.54\\
			\midrule
			Averaging (Macro)  &  51.69 & 49.59  & 49.30\\
			\textbf{Averaging (Micro)} & -- &-- & \textbf{50.62} \\
			\bottomrule
		\end{tabular}
		\vskip -0.1in
	\end{table}

	\begin{table}[ht]
		\caption{Model Performance on Mood Prediction}
		\label{tab:kmeans}
		\centering
		\begin{tabular}{ c c c  c  }
			\toprule
			\textbf{Mood Category} & \textbf{Precision} & \textbf{Recall} & \textbf{F1-Score}   \\
			\midrule
			Good Mood   & 75.60 & 73.80  & 74.68 \\
			Bad Mood&   68.57 & 70.58    & 69.56     \\
			\bottomrule
		\end{tabular}
		\vskip -0.1in
	\end{table}

	\begin{table*}[t]
		\caption{Ablation Study}
		\label{tab:ablation}
		
		\centering
		\begin{tabular}{ c|c|c|c|c|c|c|c|c|c }
			\toprule
			\multirow{2}{*}{\thead{Model}}                & 	\multirow{2}{*}{\thead{Character\\ embedding}} & 	\multirow{2}{*}{\thead{Phonetic \\ Hashing}} & 	\multirow{2}{*}{\thead{Attention}}  & 	\multirow{2}{*}{\thead{CRF}} & 	\multirow{2}{*}{\thead{Model Size}} &  
			\multicolumn{3}{c|}{\thead{DailyDialog}} & \multirow{2}{*}{\thead{Custom Data \\ F1 Score} }\\ \cline{7-9}
			& & & & & & \thead{Precision} & \thead{Recall} & \thead{F1 Score} & 
			\\ \midrule
			Model1  & \xmark & \xmark & \xmark & \xmark & 1.319 MB  &  46.92   & 53.74   &  50.06       & 42.52       \\
			Model2   &  \cmark	 & \xmark & \xmark  & \xmark & 1.676 MB   &   48.25  &  54.98 & 51.39      &    44.27       \\
			Model3  & \cmark  & \cmark & \xmark & \xmark  &   1.676 MB &  48.76   & 55.31    & 51.82   & 46.88      \\
			Model4  & \cmark & \cmark & \cmark & \xmark & 1.677 MB  &   53.63    & 58.31  &   55.87   & 47.19      \\
			\rowcolor{Gray}
			Model5 (LEAPMood)& \cmark & \cmark & \cmark  & \cmark & 1.678 MB & \textbf{59.75} & \textbf{64.55} & \textbf{62.05} & \textbf{50.62}\\ \bottomrule
		\end{tabular}
		\vskip -0.1in
	\end{table*}

	\section{Results}
	We experiment with our proposed pipeline on-device using the Samsung A50 smartphone model (6GB RAM, 64GB ROM, Samsung Exynos 7 Octa 9610).
	
	We analyse the performance of the ERC module explained in Section~\ref{s:ERC} on the DailyDialog dataset \cite{li2017dailydialog} as shown in Table \ref{tab:dailydialog}. To do this, we train our model on the DailyDialog dataset and measure the Precision, Recall, and F1 scores for each category following which we measure the Macro and Micro Averages. Since this dataset has a large number of samples (>80\%) belonging to the `other' class, we do not consider this category to calculate the Micro and Macro Averages as this trend has been followed by all the previous State-of-Art (SOA) works \cite{lee2021graph}\cite{wang2020contextualized}. The results show that our model performs exceptionally, achieving a Micro F1 Score of 62.05\%. This is more impressive considering that it has a memory footprint of a mere 1.678MB. 
	
	We compare our model with some of the recent baseline models and on the current SOA CESTa\cite{wang2020contextualized} as shown in Table~\ref{tab:soa}. We observe that our performance is very closely comparable to the SOA and stands second, behind by just 1.07\% in the Micro F1 Score of CESTa. This is with a significant reduction in model size ($\sim$90\%), as CESTa uses a 12-layer 2048 inner-layer dimension and 100 hidden dimensions along with an embedding size of 300, resulting in a model size greater than 200MB, while our model is of just 1.678MB. This is a significant contribution towards the task of ERC.
	
	Further, we test our model on our custom data explained in Section \ref{s:dataset} to evaluate its performance on real-world chat scenarios and observe that our model performs well on the basis that these contain acronyms, spelling mistakes, etc. and has more categories of emotions as shown in Table~\ref{tab:custom data}. This is possible due to the phonetic hashing component in the model the effect of which is explained in Section~\ref{s:ablation}.

	\begin{figure}[t]
		\centering
		\includegraphics[width=0.9\linewidth]{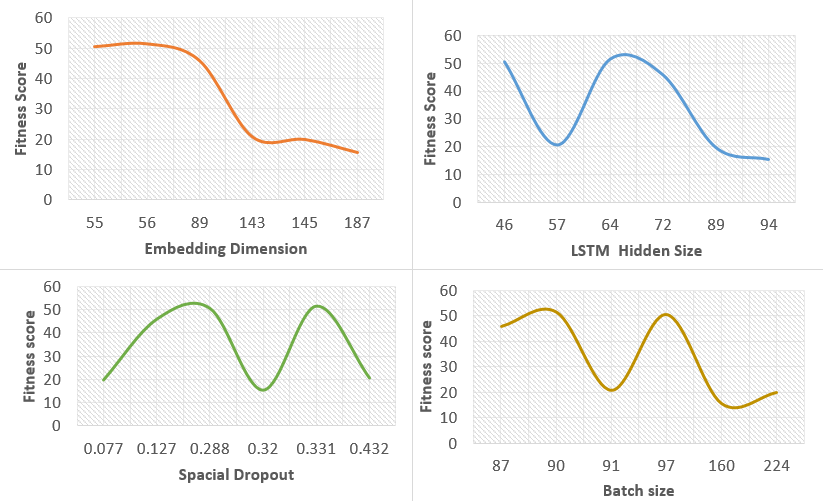}
		\caption{Graph showing Different  Hyper parameter values vs Fitness Function during the process of tuning using GA}
		\label{fig:graph}
		\vskip -0.1in
	\end{figure}
	
	We study the effect of using Hyperparameter tuning fine-tuned for on-device deployment as explained in Section~\ref{s:HP}. Fig.~\ref{fig:graph} shows the variation of some of the Hyperparameters as a function of Fitness Score as observed during the optimization process using GA. We utilize the most optimal values of hyperparameters obtained after GA. We observe that GA contributes to an increase in Accuracy of 1.08\%  with a decrease in model size of 0.46MB.
	
	We evaluate the mood prediction performance of LEAPMood on the curated and labeled data as shown in Table~\ref{tab:kmeans}. We observe good performance with a Macro F1-Score of 72.12\%. More details into the model performance are observed after Error Analysis which is explained in Section~\ref{s:error}. 
	\subsection{Error Analysis} \label{s:error}
	We conduct an error analysis to observe any trends among the false predictions by our model. On analyzing ERC results, we observe that most of the errors are due to predictions of similar emotion categories. For Ex: Happy and Surprise (Wow that's great), Sadness and Fear (I cannot do this). This is because the same sentence can be part of two different labels depending on the tone and we lack tone features in the textual data. We also observe the majority of the false positives are due to the `other' class in DailyDialog.
	
	On examining the mood prediction module results, we can conclude that most of the errors are due to samples where the emotion in the early stages of the grouped data is lost slowly with time. Moods are more affected by the recent emotions compared to those felt long before which is not captured by our current model.
	
	We aim to address these as future work of this proposal.

	\subsection{Ablation Study}\label{s:ablation}
	We perform an ablation study to analyze the effect of each layer on the model performance. We eliminate each layer and study the impact on overall model results as shown in Table~\ref{tab:ablation}. The results prove that each component in the proposed architecture is crucial to obtain better performance. 
	
	The character embedding layer helps learn fine character level features and improve performance by 1.33\% in DailyDialog and 1.75\% on our custom data. Although Phonetic hashing shows a minute change on the DailyDialog dataset, it shows a significant improvement on the custom data showing that for real-world use cases, it is a good addition to improve performance. The addition of attention shows good improvement on both datasets, aligning with our expectations. Finally, the CRF layer proves to be a major contributor towards the model performance, especially on the DailyDialog dataset where the context of previous and neighboring emotions play a major role to determine the emotion category.

	\section{Conclusion}
	In this paper, we propose for the first time a novel approach to predict users' moods from textual data. Since mood is not just determined by one single message, we create a pipeline to detect the overall mood resulting from various sets of inputs. This is done by first recognizing the emotion category that the individual set of text represents and further modeling these for the final output. We create our custom dataset to improve the real-time performance of the model and analyze the overall performance of the model on our custom dataset. We introduce a novel fitness function for Hyperparameter tuning using GA and show its effectiveness. We further benchmark our emotion recognition pipeline with the DailyDialog dataset and observe a performance of 62.05\% Micro F1 score which is just 1.07\% less than the SOA and with a mere 1.67MB memory footprint. We also analyze the performance on mood prediction and achieve a Macro F1 Score of 72.12\%. In the future, to improve the performance of mood prediction we aim to implement K-means using dynamic time wrapping to take temporal information of emotions into account.

	\bibliographystyle{IEEEtran}
	\bibliography{mybib_mood}

\begin{thebibliography}{10}
\providecommand{\url}[1]{#1}
\csname url@samestyle\endcsname
\providecommand{\newblock}{\relax}
\providecommand{\bibinfo}[2]{#2}
\providecommand{\BIBentrySTDinterwordspacing}{\spaceskip=0pt\relax}
\providecommand{\BIBentryALTinterwordstretchfactor}{4}
\providecommand{\BIBentryALTinterwordspacing}{\spaceskip=\fontdimen2\font plus
\BIBentryALTinterwordstretchfactor\fontdimen3\font minus
  \fontdimen4\font\relax}
\providecommand{\BIBforeignlanguage}[2]{{%
\expandafter\ifx\csname l@#1\endcsname\relax
\typeout{** WARNING: IEEEtran.bst: No hyphenation pattern has been}%
\typeout{** loaded for the language `#1'. Using the pattern for}%
\typeout{** the default language instead.}%
\else
\language=\csname l@#1\endcsname
\fi
#2}}
\providecommand{\BIBdecl}{\relax}
\BIBdecl

\bibitem{mikus2018predicting}
A.~Mikus, M.~Hoogendoorn, A.~Rocha, J.~Gama, J.~Ruwaard, and H.~Riper,
  ``Predicting short term mood developments among depressed patients using
  adherence and ecological momentary assessment data,'' \emph{Internet
  interventions}, vol.~12, pp. 105--110, 2018.

\bibitem{data}
\BIBentryALTinterwordspacing
``Global social media stats,'' {Datareportal}, accessed on 2021-07-26.
  [Online]. Available: \url{https://datareportal.com/social-media-users}
\BIBentrySTDinterwordspacing

\bibitem{salari2020prevalence}
N.~Salari, A.~Hosseinian-Far, R.~Jalali, A.~Vaisi-Raygani, S.~Rasoulpoor,
  M.~Mohammadi, S.~Rasoulpoor, and B.~Khaledi-Paveh, ``Prevalence of stress,
  anxiety, depression among the general population during the covid-19
  pandemic: a systematic review and meta-analysis,'' \emph{Globalization and
  health}, vol.~16, no.~1, pp. 1--11, 2020.

\bibitem{statistica}
\BIBentryALTinterwordspacing
``Number of smartphone users from 2016 to 2021,'' {Statista}, accessed on
  2021-07-26. [Online]. Available:
  \url{https://www.statista.com/statistics/330695/number-of-smartphone-users-worldwide/}
\BIBentrySTDinterwordspacing

\bibitem{gartner}
\BIBentryALTinterwordspacing
``Gartner highlights 10 uses for ai-powered smartphones,'' {Gartner}, accessed
  on 2021-07-26. [Online]. Available:
  \url{https://www.gartner.com/en/newsroom/press-releases/2018-03-20-gartner-highlights-10-uses-for-ai-powered-smartphones}
\BIBentrySTDinterwordspacing

\bibitem{moodvsemotion}
\BIBentryALTinterwordspacing
``Mood vs emotion: Differences and traits,'' {Paul Ekman Group}, accessed on
  2021-07-26. [Online]. Available:
  \url{https://www.paulekman.com/blog/mood-vs-emotion-difference-between-mood-emotion/}
\BIBentrySTDinterwordspacing

\bibitem{emotionasmood}
\BIBentryALTinterwordspacing
``Emotions, feelings and moods: What’s the difference?'' {Sixseconds},
  accessed on 2021-07-26. [Online]. Available:
  \url{https://www.6seconds.org/2017/05/15/emotion-feeling-mood/}
\BIBentrySTDinterwordspacing

\bibitem{thanapattheerakul2018emotion}
T.~Thanapattheerakul, K.~Mao, J.~Amoranto, and J.~H. Chan, ``Emotion in a
  century: A review of emotion recognition,'' in \emph{Proceedings of the 10th
  International Conference on Advances in Information Technology}, 2018, pp.
  1--8.

\bibitem{picard2000affective}
R.~W. Picard, \emph{Affective computing}.\hskip 1em plus 0.5em minus
  0.4em\relax MIT press, 2000.

\bibitem{10.1145/306668.306683}
\BIBentryALTinterwordspacing
J.~F. Cohn and G.~S. Katz, ``Bimodal expression of emotion by face and voice,''
  in \emph{Proceedings of the Sixth ACM International Conference on Multimedia:
  Face/Gesture Recognition and Their Applications}, ser. MULTIMEDIA '98.\hskip
  1em plus 0.5em minus 0.4em\relax New York, NY, USA: Association for Computing
  Machinery, 1998, p. 41–44. [Online]. Available:
  \url{https://doi.org/10.1145/306668.306683}
\BIBentrySTDinterwordspacing

\bibitem{840655}
L.~De~Silva and P.~C. Ng, ``Bimodal emotion recognition,'' in \emph{Proceedings
  Fourth IEEE International Conference on Automatic Face and Gesture
  Recognition (Cat. No. PR00580)}, 2000, pp. 332--335.

\bibitem{inproceedings}
L.~Devillers, L.~Lamel, and I.~Vasilescu, ``Emotion detection in task-oriented
  spoken dialogues,'' 08 2003, pp. III-- 549.

\bibitem{nguyen}
T.~Nguyen, M.~Li, I.~Bass, and I.~Sethi, ``Investigation of combining svm and
  decision tree for emotion classification.'' 01 2005, pp. 540--544.

\bibitem{499429}
T.~Yanaru, ``An emotion processing system based on fuzzy inference and
  subjective observations,'' in \emph{Proceedings 1995 Second New Zealand
  International Two-Stream Conference on Artificial Neural Networks and Expert
  Systems}, 1995, pp. 15--20.

\bibitem{wilson2004just}
T.~Wilson, J.~Wiebe, and R.~Hwa, ``Just how mad are you? finding strong and
  weak opinion clauses,'' in \emph{aaai}, vol.~4, 2004, pp. 761--769.

\bibitem{bracewell2008semi}
D.~B. Bracewell, ``Semi-automatic creation of an emotion dictionary using
  wordnet and its evaluation,'' in \emph{2008 IEEE Conference on Cybernetics
  and Intelligent Systems}.\hskip 1em plus 0.5em minus 0.4em\relax IEEE, 2008,
  pp. 1385--1389.

\bibitem{wu2006emotion}
C.-H. Wu, Z.-J. Chuang, and Y.-C. Lin, ``Emotion recognition from text using
  semantic labels and separable mixture models,'' \emph{ACM transactions on
  Asian language information processing (TALIP)}, vol.~5, no.~2, pp. 165--183,
  2006.

\bibitem{4368008}
Z.~Teng, F.~Ren, and S.~Kuroiwa, ``Emotion recognition from text based on the
  rough set theory and the support vector machines,'' in \emph{2007
  International Conference on Natural Language Processing and Knowledge
  Engineering}, 2007, pp. 36--41.

\bibitem{yang2007emotion}
C.~Yang, K.~H.-Y. Lin, and H.-H. Chen, ``Emotion classification using web blog
  corpora,'' in \emph{IEEE/WIC/ACM International Conference on Web Intelligence
  (WI'07)}.\hskip 1em plus 0.5em minus 0.4em\relax IEEE, 2007, pp. 275--278.

\bibitem{zadeh2017tensor}
A.~Zadeh, M.~Chen, S.~Poria, E.~Cambria, and L.-P. Morency, ``Tensor fusion
  network for multimodal sentiment analysis,'' \emph{arXiv preprint
  arXiv:1707.07250}, 2017.

\bibitem{majumder2019dialoguernn}
N.~Majumder, S.~Poria, D.~Hazarika, R.~Mihalcea, A.~Gelbukh, and E.~Cambria,
  ``Dialoguernn: An attentive rnn for emotion detection in conversations,'' in
  \emph{Proceedings of the AAAI Conference on Artificial Intelligence},
  vol.~33, no.~01, 2019, pp. 6818--6825.

\bibitem{shen2020dialogxl}
W.~Shen, J.~Chen, X.~Quan, and Z.~Xie, ``Dialogxl: All-in-one xlnet for
  multi-party conversation emotion recognition,'' \emph{arXiv preprint
  arXiv:2012.08695}, 2020.

\bibitem{lee2021graph}
B.~Lee and Y.~S. Choi, ``Graph based network with contextualized
  representations of turns in dialogue,'' \emph{arXiv preprint
  arXiv:2109.04008}, 2021.

\bibitem{wang2020contextualized}
Y.~Wang, J.~Zhang, J.~Ma, S.~Wang, and J.~Xiao, ``Contextualized emotion
  recognition in conversation as sequence tagging,'' in \emph{Proceedings of
  the 21th Annual Meeting of the Special Interest Group on Discourse and
  Dialogue}, 2020, pp. 186--195.

\bibitem{jaques2017multimodal}
N.~Jaques, S.~Taylor, A.~Sano, and R.~Picard, ``Multimodal autoencoder: A deep
  learning approach to filling in missing sensor data and enabling better mood
  prediction,'' in \emph{2017 Seventh International Conference on Affective
  Computing and Intelligent Interaction (ACII)}.\hskip 1em plus 0.5em minus
  0.4em\relax IEEE, 2017, pp. 202--208.

\bibitem{cho2019mood}
C.-H. Cho, T.~Lee, M.-G. Kim, H.~P. In, L.~Kim, and H.-J. Lee, ``Mood
  prediction of patients with mood disorders by machine learning using passive
  digital phenotypes based on the circadian rhythm: prospective observational
  cohort study,'' \emph{Journal of medical Internet research}, vol.~21, no.~4,
  p. e11029, 2019.

\bibitem{liu2020multimodal}
T.~Liu, P.~P. Liang, M.~Muszynski, R.~Ishii, D.~Brent, R.~Auerbach, N.~Allen,
  and L.-P. Morency, ``Multimodal privacy-preserving mood prediction from
  mobile data: A preliminary study,'' \emph{arXiv preprint arXiv:2012.02359},
  2020.

\bibitem{roshanaei2015features}
M.~Roshanaei, R.~Han, and S.~Mishra, ``Features for mood prediction in social
  media,'' in \emph{2015 IEEE/ACM International Conference on Advances in
  Social Networks Analysis and Mining (ASONAM)}.\hskip 1em plus 0.5em minus
  0.4em\relax IEEE, 2015, pp. 1580--1581.

\bibitem{falkner2018bohb}
S.~Falkner, A.~Klein, and F.~Hutter, ``Bohb: Robust and efficient
  hyperparameter optimization at scale,'' in \emph{International Conference on
  Machine Learning}.\hskip 1em plus 0.5em minus 0.4em\relax PMLR, 2018, pp.
  1437--1446.

\bibitem{akiba2019optuna}
T.~Akiba, S.~Sano, T.~Yanase, T.~Ohta, and M.~Koyama, ``Optuna: A
  next-generation hyperparameter optimization framework,'' in \emph{Proceedings
  of the 25th ACM SIGKDD international conference on knowledge discovery \&
  data mining}, 2019, pp. 2623--2631.

\bibitem{li2017dailydialog}
Y.~Li, H.~Su, X.~Shen, W.~Li, Z.~Cao, and S.~Niu, ``Dailydialog: A manually
  labelled multi-turn dialogue dataset,'' \emph{arXiv preprint
  arXiv:1710.03957}, 2017.

\bibitem{beider2010phonetic}
A.~Beider and S.~P. Morse, ``Phonetic matching: A better soundex,''
  \emph{Association of Professional Genealogists Quarterly [serial on the
  Internet]}, 2010.

\end{thebibliography}
	
\end{document}